\documentclass[letterpaper, 10 pt, conference]{ieeeconf}  % 

\IEEEoverridecommandlockouts                              % This command is only needed if 
\usepackage{cite}
\usepackage[pdftex]{graphicx}

\usepackage{pifont}

\DeclareGraphicsExtensions{.pdf,.jpeg,.png,.tif}

\usepackage{amsmath}

\title{\LARGE \bf
Compliant Suction Gripper with Seamless Deployment and Retraction for Robust Picking against Depth and Tilt Errors
}

\author{Yuna Yoo, Jaemin Eom, Min Jo Park, and Kyu-Jin Cho*
\thanks{This work was supported in part by The Ministry of Trade, Industry and Energy (MOTIE, Korea, Project Number: 20008908), and in part by the National Research Foundation of Korea (NRF) Grant funded by the Korean Government (MSIT) (NRF-2016R1A5A1938472).(Yuna Yoo and Jaemin Eom contributed equally to this work.) (Corresponding author: Kyu-Jin Cho.)}% <-this % stops a space
\thanks{Yuna Yoo, Jaemin Eom, Min Jo Park, and Kyu-Jin Cho are with the Biorobotics Laboratory, Soft Robotics Research Center, Department of Mechanical Engineering, Institute of Engineering Seoul National University, Seoul 08826, Republic of Korea
(e-mail: yunayoo@snu.ac.kr; jaemineom@snu.ac.kr; mjpark1995@gmail.com; kjcho@snu.ac.kr)}
}

\begin{document}

\maketitle
\thispagestyle{empty}
\pagestyle{empty}

%%%%%%%%%%%%%%%%%%%%%%%%%%%%%%%%%%%%%%%%%%%%%%%%%%%%%%%%%%%%%%%%%%%%%%%%%%%%%%%%
\begin{abstract}

Applying suction grippers in unstructured environments is a challenging task because of depth and tilt errors in vision systems, requiring additional costs in elaborate sensing and control. To reduce additional costs, suction grippers with compliant bodies or mechanisms have been proposed; however, their bulkiness and limited allowable error hinder their use in complex environments with large errors. Here, we propose a compact suction gripper that can pick objects over a wide range of distances and tilt angles without elaborate sensing and control. The spring-inserted gripper body deploys and conforms to distant and tilted objects until the suction cup completely seals with the object and retracts immediately after, while holding the object. This seamless deployment and retraction is enabled by connecting the gripper body and suction cup to the same vacuum source, which couples the vacuum picking and retraction of the gripper body. Experimental results validated that the proposed gripper can pick objects within 79 mm, which is 1.4 times the initial length, and can pick objects with tilt angles up to 60°. The feasibility of the gripper was verified by demonstrations, including picking objects of different heights from the same picking height and the bin picking of transparent objects.

\end{abstract}


\section{INTRODUCTION}

Suction grippers have been widely used for pick-and-place tasks due to their robustness, versatility, and high operating speed \cite{Pham2019d, Papadakis2020c, Huh2021c}. With growing demand for automatic picking systems in unstructured environments, recent studies on suction grippers focus not only on picking in highly-structured environments such as factory lines but also on picking in cluttered or unstructured real-world environments including bin picking \cite{Mueller2016a}, warehouse picking \cite{Schwarz2017}, and food handling \cite{Joffe2019c, Burks2005a}.

While identifying the exact location of an object is important for reliable picking, recognizing the accurate distance and tilt angle between the suction gripper and object remains a challenge when picking in unstructured environments. The errors in the depth and tilt data obtained from vision systems can be up to tens of centimeters and degrees, respectively, due to factors such as transparency, reflectiveness, lack of texture, black object color \cite{Sajjan2020a, Baek2020, Kim2017c, Chai2020e}, background illumination \cite{Piao2020d, Wang2021d}, or occlusion by the gripper \cite{Aoyagi2020c, Koyama2019d}. These depth and tilt errors lower the reliability of picking, as approaching at an improper distance or angle results in the failure of picking, or damages the object \cite{Baek2021c, Liu2020b}. Studies have been conducted to solve this reliability problem using computational algorithms, machine learning \cite{Cui2013a, Chang2021a}, or by using vision systems in combination with tactile sensors to measure and correct the depth and tilt errors \cite{Aoyagi2020c, Doi2020c}. However, additional costs are incurred for additional sensor and control systems or complex algorithms.

To reduce vision, sensing, or control costs, grippers that can offload system complexities onto intelligent mechanical designs based on ‘physical intelligence’ have been developed \cite{Sitti2021d, Laschi2016a}. Grippers with physical intelligence can improve the reliability of picking by conforming to various shapes and sizes of objects \cite{Li2019, Amend2012g, Zhou2017c} or misaligned objects \cite{Li2019, Pounds2011d, Angelini2020d} through the compliance of the end effector. To alleviate the depth errors with physical intelligence, vacuum cylinders can be used \cite{Vac-cyl}, which operate by pushing the suction cup forward and pulling it back when a suction seal is formed. However, vacuum cylinders cannot compensate for tilt errors due to the rigidity of the cylinders, and because they occupy several times the space required for their stroke when at rest, rendering the system bulky. On the other hand, bellows suction cups can mitigate both depth and tilt errors, because their bellows-shaped body conforms to the object by folding in response to object contact \cite{Liu2020, McCulloch2016}. However, the allowable depth error, which is determined by the length of the bellows-shaped body, is difficult to be designed for more than a few centimeters due to the risk of uncontracted bellows-shaped body colliding with environment.

In this letter, we developed a compliant deployable suction gripper with a compact structure that enables reliable picking against a wide range of depth and tilt errors, without additional sensing and control. The proposed gripper longitudinally deploys its compliant body while conforming to a distant or tilted object, and retracts while holding the object as soon as the suction cup seals against the object as shown in Fig. 1. The gripper remains contracted after releasing the object to be in a compact form when not in operation. There are two key enabling technologies for gripper motion. The deployable and bendable spring-inserted gripper body enables conforming to distant and tilted objects. Connecting the gripper body and suction cup to the same vacuum source enables seamless deployment and retraction by coupling the vacuum picking and retraction of the gripper body. With these enabling technologies, the proposed gripper compensates for depth and tilt errors up to 79mm and 60 degrees which are six and two times larger, respectively, compared to the bellows suction cup of similar length to the proposed gripper. The proposed gripper picks up objects within 1.2 seconds of cycle time due to fast and seamless deployment and retraction, and picks up objects weighing up to 4400g, also having robustness verified by 1000 cycles of repeatability test.  The picking ability of the proposed gripper against depth and tilt errors was verified by demonstrations, including picking objects with different heights from the same picking height and bin picking of transparent objects. Furthermore, the gripper was applied to depalletizing tasks in which the number of grippers does not match the number of objects arranged on each layer. Additionally, a gripper with extended stroke of 140mm was applied to pick objects in narrow, deep spaces, which cannot be reached by manipulators.

   \begin{figure}[t]
      \centering
      \includegraphics[width=0.97\columnwidth]{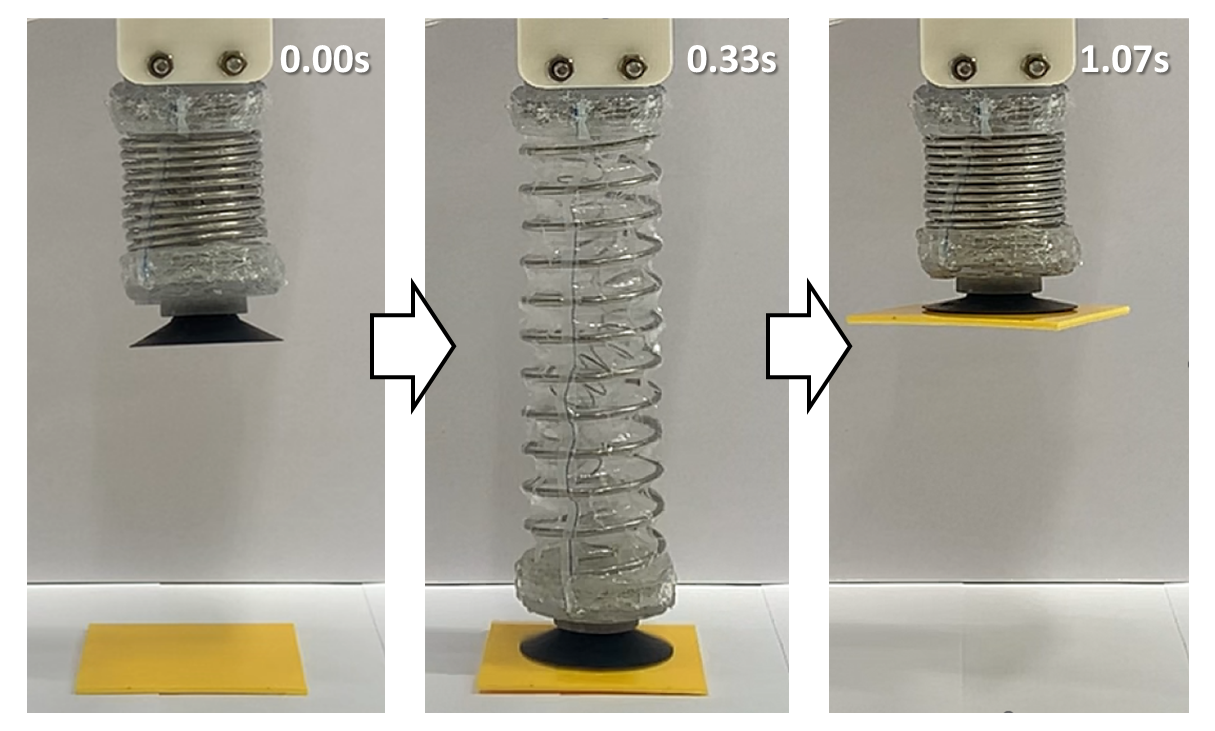}
      \caption{Proposed suction gripper picking up a distant object through seamless deployment and retraction.}
      \label{fig_1}
      \vspace{-0.5cm}
   \end{figure}

The rest of this letter is organized as follows: Section II describes two major designs: pneumatic line design and the mechanical design of the gripper. Section III presents the theoretical modeling of the allowable depth errors and picking force of the gripper. Section IV presents the experimental verification of the performance of the gripper. Section V presents the demonstrations of the proposed gripper. Section VI concludes the letter.

\section{Mechanical Design and Fabrication}
   
\subsection{Overall Structure and Picking Strategy}

    \begin{figure*}[t]
      \centering
      \includegraphics[width=0.97\textwidth]{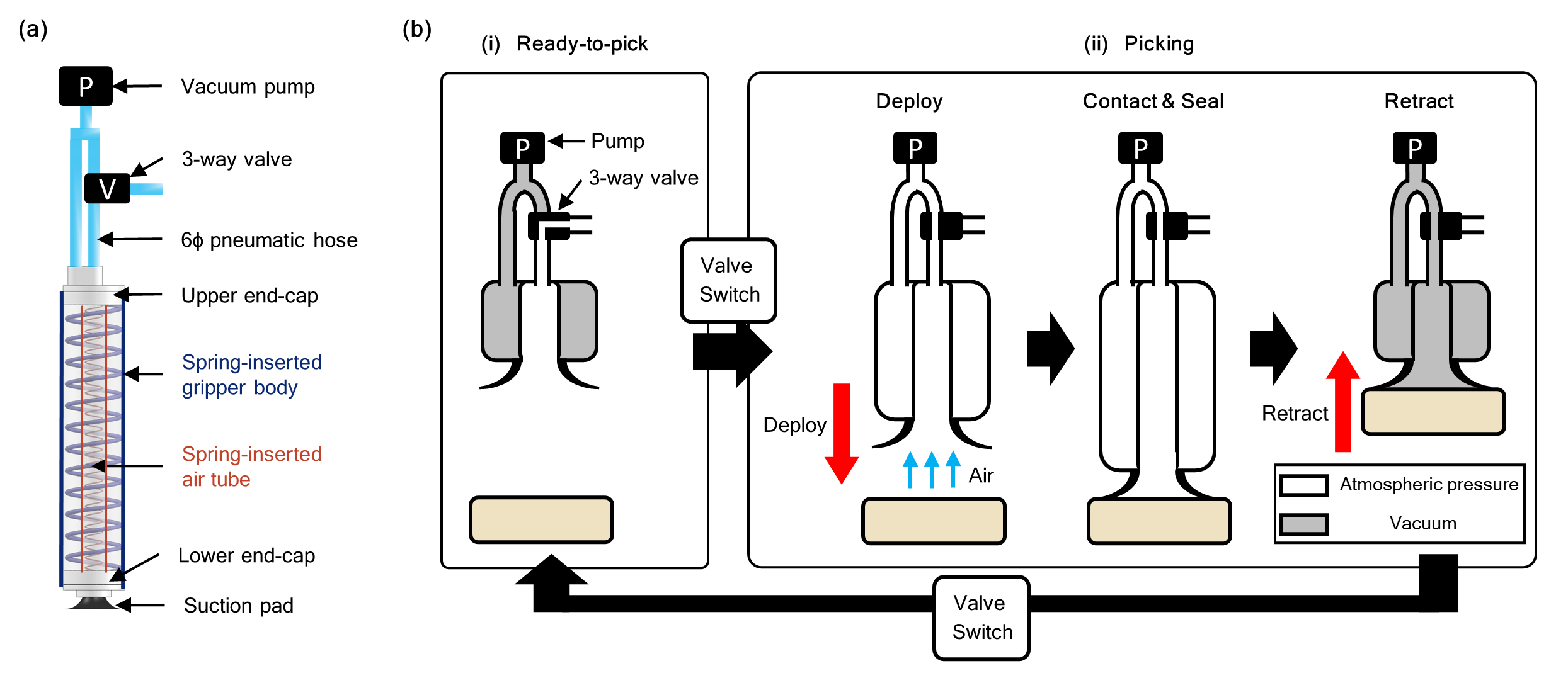}
      \caption{Design and strategy of the proposed gripper. (a) Overall structure of the proposed gripper system: the system comprises a gripper containing two concentric chambers and a pneumatic circuit connected to the gripper. (b) Picking strategy of the gripper: (i) the gripper is contracted in the ready-to-pick stage; (ii) the gripper moves on to the picking stage when the valve switches, and seamlessly deploys and retracts to pick up an object. }
      \label{fig_2}
      \vspace{-0.5cm}
   \end{figure*}
   
To pick up objects against depth and tilt errors, the proposed gripper was designed to deploy longitudinally and retract while holding the object as soon as the gripper seals against the object. The key design requirement for this gripper operation was a system that enables seamless deployment and retraction without sensors, which was facilitated by structurally embedding the operation sequence into the proposed gripper and pneumatic circuit.

The overall structure of the gripper system is shown in Fig. 2(a). The proposed gripper comprised two concentric cylindrical chambers, where the outer and inner chambers are termed as the spring-inserted gripper body and spring-inserted air tube, respectively. The gripper body retracts and deploys when negative pressure is applied or released, and the air tube provides the airway to the suction cup. The pneumatic circuit included a vacuum pump (N035AN.18, KNF Neuberger Inc.) and a 3-way solenoid valve (UV307-4LL, Shinyeong Mechatronics Co., Ltd.). In the circuit, the gripper body was directly connected to the vacuum pump and the air tube was connected to the pump or atmosphere through the 3-way valve.

The picking strategy utilizing a pneumatic circuit and deployable gripper is illustrated in Fig. 2 (b). First, in the ‘ready-to-pick’ stage, the valve connects the suction cup and the atmosphere. The gripper body becomes a closed space connected to the vacuum pump; therefore, the gripper body is vacuumized by the pump and contracted. The gripper moves to the picking stage when the valve switches and connects the suction cup and gripper body. Thereafter, the gripper body is connected to the atmosphere through the suction cup, so the sealing is released and the body deploys. When the suction cup contacts and seals with the object, the entire space inside the gripper is vacuumized, and the gripper body retracts immediately, while the suction cup holds the object. Finally, when the valve switches and connects the suction cup to the atmosphere, the gripper returns to the ready-to-pick stage. The suction cup places the object down, and the gripper body remains contracted. Connecting the suction cup and gripper body to the same vacuum source throughout the picking stage enables seamless deployment and retraction without sensors, and disconnecting the suction cup and gripper body when releasing the object enables the gripper to contract when not in use.

\subsection{Gripper Design}
The gripper body was designed as a compressible spring enclosed in a cylindrical LDPE film for the following reasons: First, the gripper can be deployed at a high speed due to the restoring force of the spring. Second, the spring is bendable; therefore, the gripper can conform to tilted objects during deployment. Finally, the longitudinally compressible and radially incompressible characteristics of the spring allow the gripper to retract effectively in the longitudinal direction. 

The spring-inserted air tube should retain the airway even when the gripper retracts, and should not interfere with the operation of the gripper. When this airway is connected outside the gripper body through a pneumatic hose, the hose can disturb the gripper operation in unstructured environments. Therefore, in the proposed gripper, the airway was built inside the gripper body with a compressible spring encased in a cylindrical LDPE film with the same structure as the gripper body. The radial incompressibility of the spring prevents airway clogging during the retraction. To prevent the air tube from affecting the movement of the gripper body, the wire diameter of the spring in the air tube was determined to be 0.5 mm, which is the thinnest to be manufactured. 

   \begin{figure}[t]
      \centering
      \includegraphics[width=0.97\columnwidth]{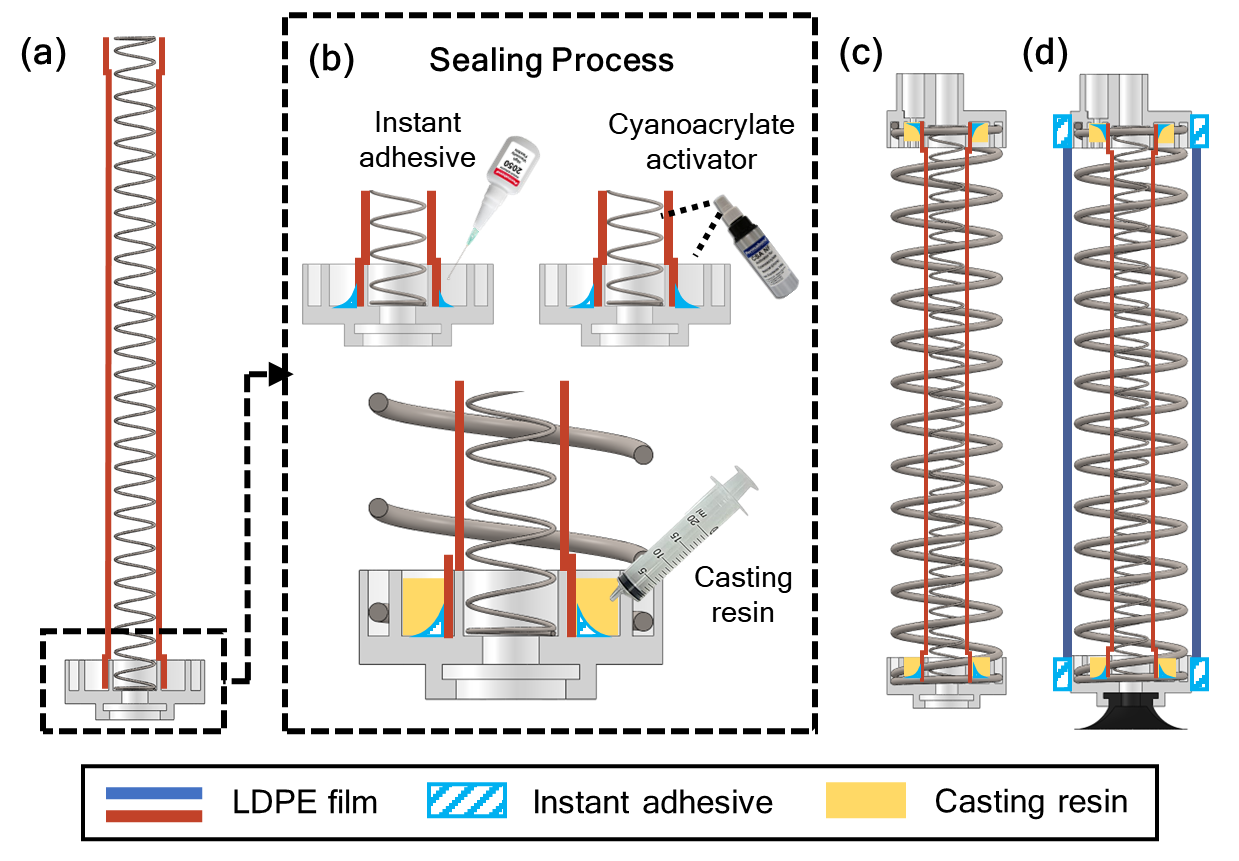}
      \caption{Fabrication process of the proposed gripper: (a) Assembling the inner spring, inner LDPE film, and the lower end-cap; (b) sealing process using instant adhesive and casting resin; (c) assembling the outer spring and the upper end-cap; (d) sealing the outer LDPE film and plugging in the suction cup.}
      \label{fig_3}
      \vspace{-0.5cm}
   \end{figure}

\subsection{Gripper Fabrication}
The upper and lower end-caps were 3D printed using OBJET30 Dental Prime (Stratasys Ltd.). The outer and inner LDPE films are fabricated by heat-pressing two 60 $\mu m$ thick planar LDPE sheets at 230 $^{\circ}$C. Fig. 3 shows the overall fabrication process of the gripper. First, the inner spring was inserted into the inner LDPE film. The inner LDPE film and the lower end-cap were sealed with instant adhesive (Permabond 2050, Permabond Engineering Adhesives Ltd.), and a cyanoacrylate activator (Permabond CSA-NF, Permabond Engineering Adhesives Ltd.) was used to promote a faster cure. Additional sealing was performed using casting resin (TASK™ 2, Smooth-on Inc.) to prevent possible air leak from the sealed components. The casting resin was poured into an end-cap and cured at room temperature for an hour. Subsequently, the outer spring was inserted, and the assembled parts were covered and sealed with an upper end-cap. The assembled parts were placed in the outer LDPE film, which was sealed with the upper and lower end-caps using the same sealing process. Finally, the suction cup was placed at the connection part of the lower end-cap.
   
\section{Modeling}
To analyze the picking ability of the proposed gripper according to the depth error, two analytical models were devised: the picking distance range from objects for successful picking and the maximum picking force according to the distance from the object.

\subsection{Picking distance range}

To analyze the allowable depth error, that is, the picking distance range for the gripper to successfully pick up an object, the maximum and minimum distances were modeled. The distance was defined as the distance between the upper surface of the upper end-cap and the upper surface of the object.

The minimum distance ($d_{min}$) is the maximum contracted length of the gripper. Two cases of minimum distance exist: the gripper contracts until the restoring force of the spring and contracting force caused by the vacuum are in equilibrium (1); the gripper contracts completely when the maximum restoring force of the spring is weaker than the contracting force (2).
\begin{equation}\label{(1)}
d_{min}=L-\tfrac{P_{atm}A_b}{k_b}+h_{endcap}+h_{suction}
\end{equation}
\begin{equation}\label{(2)}
d_{min}=nt_{spring}+2(n-1)t_{film}+h_{endcap}+h_{suction}
\end{equation}
where $h_{endcap}$ is the sum of the heights of the two end-caps, $h_{suction}$ is the height of the suction cup, $t_{spring}$ is the wire diameter of the spring, and $t_{film}$ is the LDPE film thickness.

The maximum distance ($d_{max}$) was defined as the fully deployed length of the gripper. Fig. 4 shows the schematic of the fully deployed gripper at the steady state. A pressure drop occurs from $P_{atm}$ to $P_b$ along the spring-inserted air tube and polyurethane (PU) tube \textcircled{\scriptsize 1} due to the airflow, whereas the pressure inside the gripper body and the PU tube \textcircled{\scriptsize 2} are constant with $P_b$  because of the lack of airflow along this path. The fully deployed length is smaller than the initial length, L, due to the contraction caused by the pressure difference ($\Delta P$) between the inside ($P_b$) and outside ($P_{atm}$) of the gripper body.

\begin{figure}[t]
      \centering
      \includegraphics[width=0.97\columnwidth]{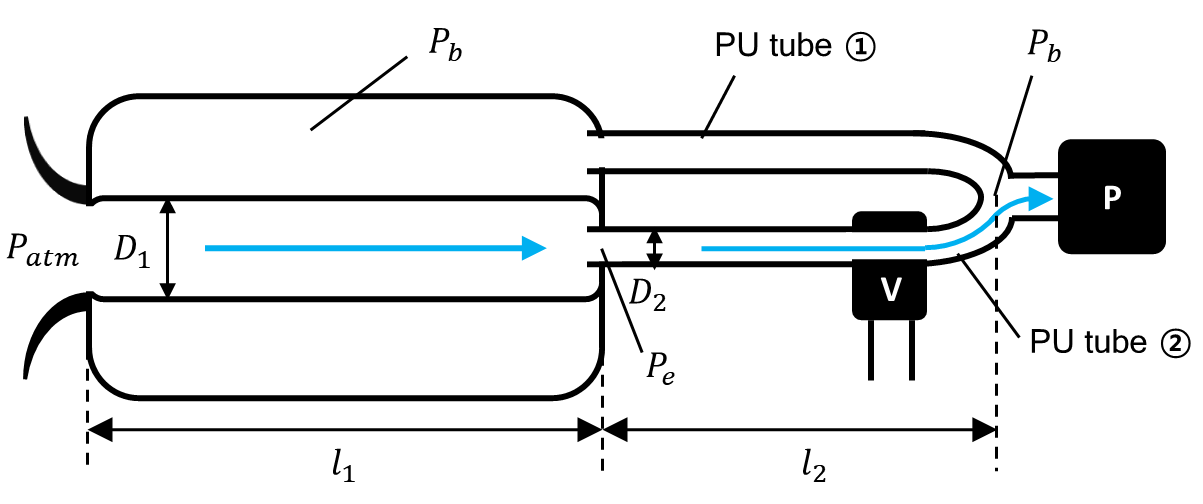}
      \caption{Schematic of the gripper system at the fully-deployed steady-state. }
      \label{fig_4}
      \vspace{-0.5cm}
   \end{figure}

Assuming that the air is incompressible and that the effects of gravity and minor losses due to the elbow, fittings, and valves are negligible, the pressure drop due to airflow is generally derived as follows (Equation (6.10) in \cite{FluidMech}):
\begin{equation}\label{(3)}
\frac{\Delta P}{\rho g}=h_f=\frac{8fL_tQ^2}{\pi ^2gD^5}
\end{equation}
where $\rho$ is density of air, $h_f$ is head loss, $f$ is friction factor of the air tube, $L_t$ is the length of tube, $Q$ is the flow rate of air, and $D$ is the tube diameter. The pressure drop along the spring-inserted air tube was ignored, because it was approximately 200 times smaller than the pressure drop along the PU tube based on (3), due to the difference between the diameter and length. From the minimum Reynolds number in the range of Q, which varies from 4–14 L/min, the friction factor can be obtained as follows:
\begin{equation}\label{(4)}
Re_{D,min}=\tfrac{\rho VD_2}{\mu }=\tfrac{4\rho Q_{min}}{\mu \pi D_2}=4453.3
\end{equation}
\begin{equation}\label{(5)}
\tfrac{1}{f^{1/2}}\approx -1.8log\left [ \tfrac{6.9}{Re_D}+\left ( \tfrac{\epsilon /D_2}{3.7} \right )^{1.11} \right ]
\end{equation}
where $V$ is airflow speed, $\mu$ is the specific gravity of air, and $\epsilon$ is the roughness value of the tube. By substituting (4) and (5) into (3), $\delta P$ is obtained and $d_{max}$ can be derived as follows:
\begin{equation}\label{(6)}
d_{max}=L-\tfrac{\Delta PA_b}{k_b}+h_{endcap}+h_{suction}
\end{equation}

\subsection{Maximum Picking Force}
   
    \begin{figure*}[t]
      \centering
      \includegraphics[width=0.97\textwidth]{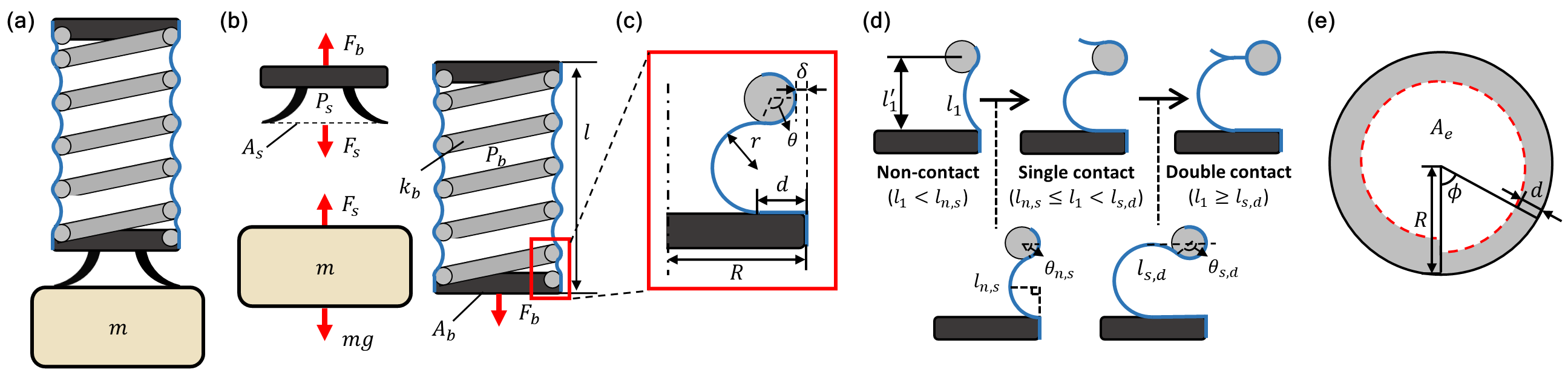}
      \caption{(a) Configuration of the gripper body when the suction cup is in contact with an object. (b) Free body diagram of the gripper body, suction cup, and the object. (c) Configuration of the film that contacts the lower end-cap as the gripper body retracts. (d) Three states of film configurations during retraction: non-contact, single contact, and double contact states. (e) Top-view of lower end-cap while film contacts to the end-cap: the gray area indicates the contact area.}
      \label{fig_5}
      \vspace{-0.5cm}
   \end{figure*}

The maximum picking force exerted by the gripper according to the distance from the object was analyzed. The configuration and free-body diagram of the gripper body and suction cup when it is in contact with the object are shown in Fig. 5 (a) and (b). We assumed that the weights of the spring and other components of the gripper were negligible throughout the picking process. In this paper, the force applied by the gripper body to lift the object is called the lifting force $F_b$, and that applied by the suction cup on the object is called the holding force $F_s$. The lifting and holding forces satisfy the following equation and inequality:
\begin{equation}\label{(7)}
F_b = (P_{atm}-P_b)A_b - k_b(L-l)
\end{equation}
\begin{equation}\label{(8)}
F_s\leq (P_{atm}-P_s)A_s
\end{equation}
where $P_{atm}$ is the atmospheric pressure; $P_b$, $P_s$, $A_b$, and $A_s$ are the internal pressure and effective area of the gripper body and suction cup, respectively; $k_b$ is the coefficient of spring; and $L$ and $l$ are the initial and current lengths of the spring, respectively.

For the gripper to lift an object without dropping it, the following condition should be satisfied:
\begin{equation}\label{(9)}
min(F_{b,max},F_{s,max})\geq mg
\end{equation}
where $m$ is the weight of the object, $F_{s,max}$ and $F_{b,max}$ are the forces when the internal pressure is closest to the vacuum. This is because the gripper cannot retract or drop the object when the weight of the object exceeds the minimum of $F_{s,max}$  and $F_{b,max}$, which is therefore the maximum picking force of the gripper. 

$F_{s,max}$ and $F_{b,max}$ have linear relationship with the effective areas of the suction cup and gripper body, respectively, as expressed in (7) and (8). The effective area of the suction cup is constant regardless of the distance from the object, however, that of the gripper body decreases as the gripper approaches the object. This is because the LDPE film slacks and sticks to the upper and lower end-caps during gripper retraction (Fig. 5(c)) \cite{Lee2019a}. Therefore, to analyze the maximum picking force, the effective area of the gripper body was modeled according to the distance between the suction cup and object. 

The effective area model was based on one fundamental condition called the contact condition: the film contacted tangentially to the upper and lower end-caps and spring. According to the initial length of the film ($l_1$), there are three film configurations (Fig. 5(d)): non-contact, single contact, and double contact. The non-contact state indicates that the film and lower end-cap lacked contact; the single contact state means that the film and lower end-cap are in contact; and the double contact state means that the film completely wraps around the spring and they are in contact with each other. The boundary value of $l_1$ between states is derived as follows:
\begin{equation}\label{(10)}
l_{n,s}=\left ( \left ( \tfrac{{l'_1}^2+t+1}{2l'_1} \right )\left ( \theta _{n,s}+\tfrac{\pi }{2} \right ) \right )+\frac{t\theta _{n.s}}{2}
\end{equation}
\begin{align}\label{(11)}
    \begin{split}
    l_{s,d}=\tfrac{1}{2}( t\left ( \theta _{s,d}+1+sin\left ( \theta _{s,d}-\tfrac{\pi }{2} \right ) \right ) \\
    +  l'_1\left ( \theta _{s,d}+\tfrac{\pi}{2}+sin\left ( \theta _{s,d}-\tfrac{\pi}{2} \right ) \right )+1
    \end{split}
\end{align}
where $l_{n,s}$ and $l_{s,d}$ are the boundary values between the non-contact and single contact states, and between the single and the double contact states, respectively; $l'_1$ is the distance between the cross section of the spring and lower end-cap; $\theta_{n,s}$ and $\theta_{s,d}$ are the angles at which the film wraps around the spring and are derived as follows:
\begin{equation}\label{(12)}
\theta_{n,s}=cos^{-1}\left ( \tfrac{(t_{spring}/2)+1}{\left ( ({l'_1}^2+t+1)/2l'_1 \right )+(t_{spring}/2)} \right ) 
\end{equation}
\begin{equation}\label{(13)}
\theta_{s,d}=\pi -sin^{-1}\left ( \frac{l'_1}{l'_1+t_{spring}} \right ) 
\end{equation}
The length of the film in contact with the lower end-cap, which is called the contact distance ($d$) in this paper, is derived using the following equations:
\begin{equation}\label{(14)}
\left\{\begin{matrix} r+\left ( \left ( r+\tfrac{t}{2} \right )sin\theta  \right )-l'_1=0
 \\r\left ( \theta +\tfrac{\pi}{2} \right )+\tfrac{t}{2}\theta
+d-l_1=0 \\\delta+\tfrac{t}{2}-\left ( r+\tfrac{t}{2} \right )cos\theta -d=0
\end{matrix}\right.
\end{equation}
\begin{equation}\label{(15)}
d=\tfrac{1}{2}\left ( l_1-\tfrac{\pi}{2}l'_1-\left ( \tfrac{\pi}{2}-1 \right )t+\delta \right )
\end{equation}
where $r$ is the radius of the non-contact part of the film, $\theta$ is the wrapped angle by film, and $\delta$ is the difference between the radius of the outer spring and the end-cap. For the single and double contact states, $d$ is derived through (14) and (15), respectively.

Because the spring is spirally formed, $l_1$, $l'_1$, and d vary along the circumference of the end-cap (Fig. 5(e)). Therefore, the effective area is obtained as follow:
\begin{equation}\label{(16)}
A_e=\int_{0}^{2\pi}\frac{1}{2}\left ( R-d \right )^2d\phi 
\end{equation}
where $A_e$ is the effective area and $R$ is the radius of the lower end-cap.

\section{Experiments}

\subsection{Picking Range}
The picking range comprises the picking distance and picking angle ranges, corresponding to the allowable depth and tilt error, respectively. First, the picking distance range according to the flow rate and spring constant of the gripper was measured and compared with the modeling to validate the robustness of the proposed gripper against depth error. Four grippers with different outer-spring constants were prepared, each with wire diameters of 1.4, 1.6, 1.8, and 2.0 mm. The other parameters of the spring, including spring constant k, are listed in Table 1.

\begin{table}[h]
\caption{Spring Parameters}
\vspace{-0.3cm}
\label{table_example}
\begin{center}
\begin{tabular}{|c|c|c|c|c|c|}
\hline
& \multicolumn{4}{c|}{Outer Spring} & Inner Spring\\
\hline
Free Length ($L$,mm) & \multicolumn{4}{c|}{150} & 150\\
\hline
Radius ($R_s$,mm) & \multicolumn{4}{c|}{15} & 5\\
\hline
Pitch ($p$,mm) & \multicolumn{4}{c|}{11.5} & 5.36\\
\hline
Wire Diameter ($t$, mm) & 1.4 & 1.6 & 1.8 & 2.0 & 0.5 \\
\hline
Spring Constant ($k$, N/m) & 125 & 198 & 326 & 522 & 22.3 \\
\hline
\end{tabular}
\end{center}
\vspace{-0.3cm}
\end{table}

The picking distance range and deployment ratio, which were calculated by dividing the fully deployed length by the contracted length, are shown in Fig. 6. As the wire diameter of the spring increased, both the contracted and fully deployed lengths increased, because a thicker spring applies a larger restoring force. In addition, as the flow rate increased, the fully deployed length decreased due to the pressure drop in the pneumatic line, as analyzed in Section III; therefore, the deployment ratio decreased. Irrespective of the spring constant and flow rate, the deployment ratio of the gripper was within 1.5–3, due to the compact configuration when not in operation. The analytical model and experimental results of the maximum distance agreed within a 7 mm error. However, the analytically predicted minimum distance was smaller than the experimental results by up to 25 mm. This difference is because the proposed model could not consider the effects of spring-inserted air tube and film wrinkling, which prevents the gripper body from contracting to maximum.

To validate the robustness of the gripper against tilt errors, the success rate of picking up a 5 g plate lying on the tilted structure for varying distance and tilt angle values was measured. The distance to the object ($l_{dist}$) is defined as the distance between the upper surface of the gripper and the center of the contact point, and the tilt angle ($\theta_{tilt}$) is the angle of the tilted structure from the horizontal line. Similarly as in the previous experiment, four grippers of wire diameters 1.4, 1.6, 1.8, and 2.0 mm were prepared, and the picking was repeated 10 times for each distance and tilt angle.

    \begin{figure}[t]
      \centering
      \includegraphics[width=0.97\columnwidth]{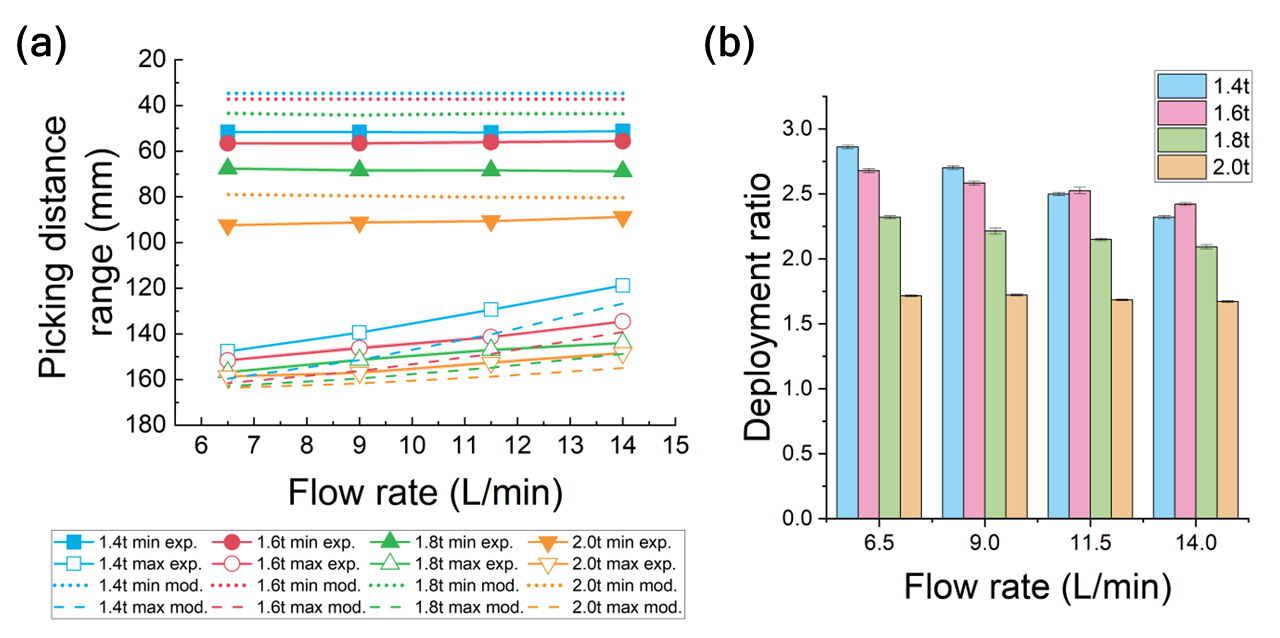}
      \caption{Experimental verification for the picking distance range. (a) Maximum and minimum distance and (b) deployment ratio of the gripper according to the flow rate and the diameter of spring wire.}
      \label{fig_6}
      \vspace{-0.5cm}
    \end{figure}
    \begin{figure*}[t]
      \centering
      \includegraphics[width=0.97\textwidth]{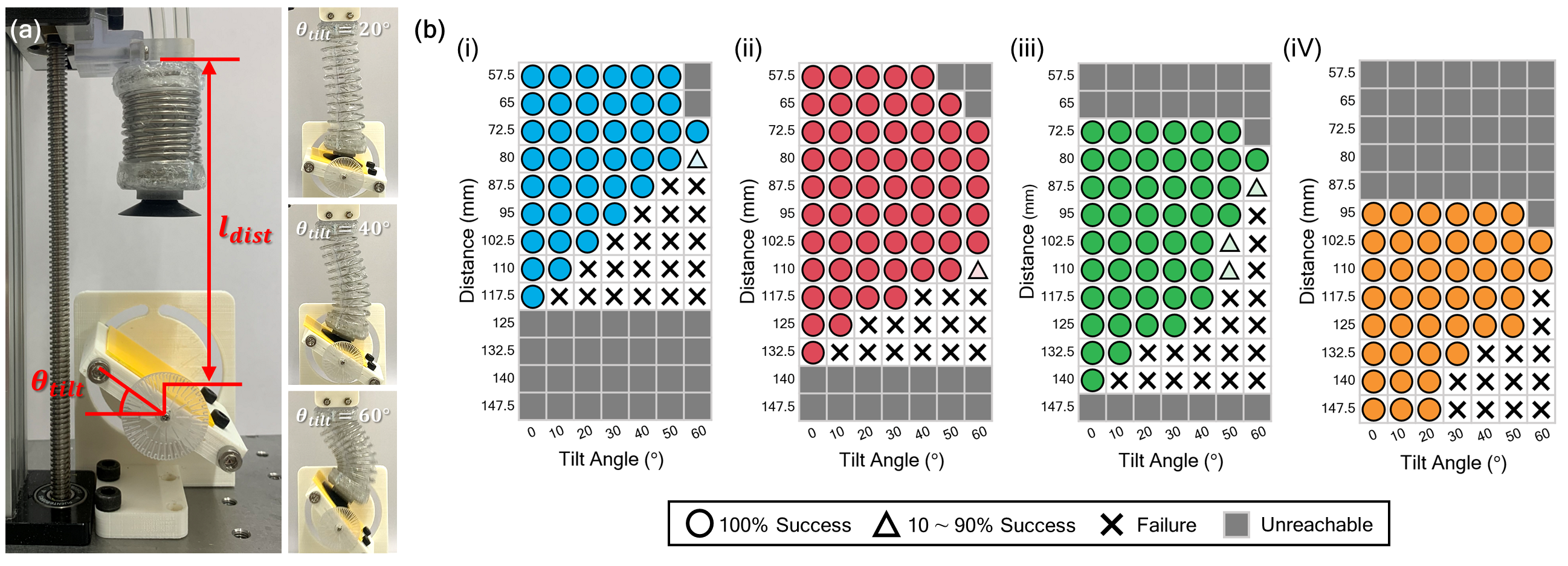}
      \caption{Picking range experiment. (a) Experimental setup for the picking range test. (b) Success rate for picking distant and tilted object with the gripper having (i) 1.4mm (ii) 1.6mm (iii) 1.8mm (iV) 2.0mm of wire diameter. }
      \label{fig_7}
      \vspace{-0.5cm}
    \end{figure*}
As shown in Fig. 7, for all four grippers, as the distance increased, failure occurred at increasingly smaller tilt angles because of the two failure modes. The first failure mode primarily occurred for grippers with small spring constants, where the suction cup could not reach the plate because of the insufficient restoring force of the gripper for conforming to the object. The second failure mode, primarily occurred for grippers with large spring constants, where slip occurred because of the insufficient friction force between the suction cup and plate for stopping the rapidly deploying gripper. Consequently, the gripper of wire diameter 1.6 mm showed the widest picking range. For $\theta_{tilt}$ larger than $60^{\circ}$, picking failed for all grippers under the second failure mode because of insufficient normal force to generate sufficient friction.

Fig. 8 summarizes the picking ability against the depth and tilt errors of the proposed gripper, and compares it with the flat and bellows suction cups. The allowable depth and tilt error range of the flat and bellows suction cups were measured using the same experimental setup used for the proposed gripper. The allowable depth and tilt error of the gripper were 140 mm and $60^{\circ}$, respectively, which are two to 26 times larger than those of the flat or bellows suction cups, verifying the applicability of the proposed gripper.

\subsection{Picking Force}

    \begin{figure}[t]
      \centering
      \includegraphics[width=0.97\columnwidth]{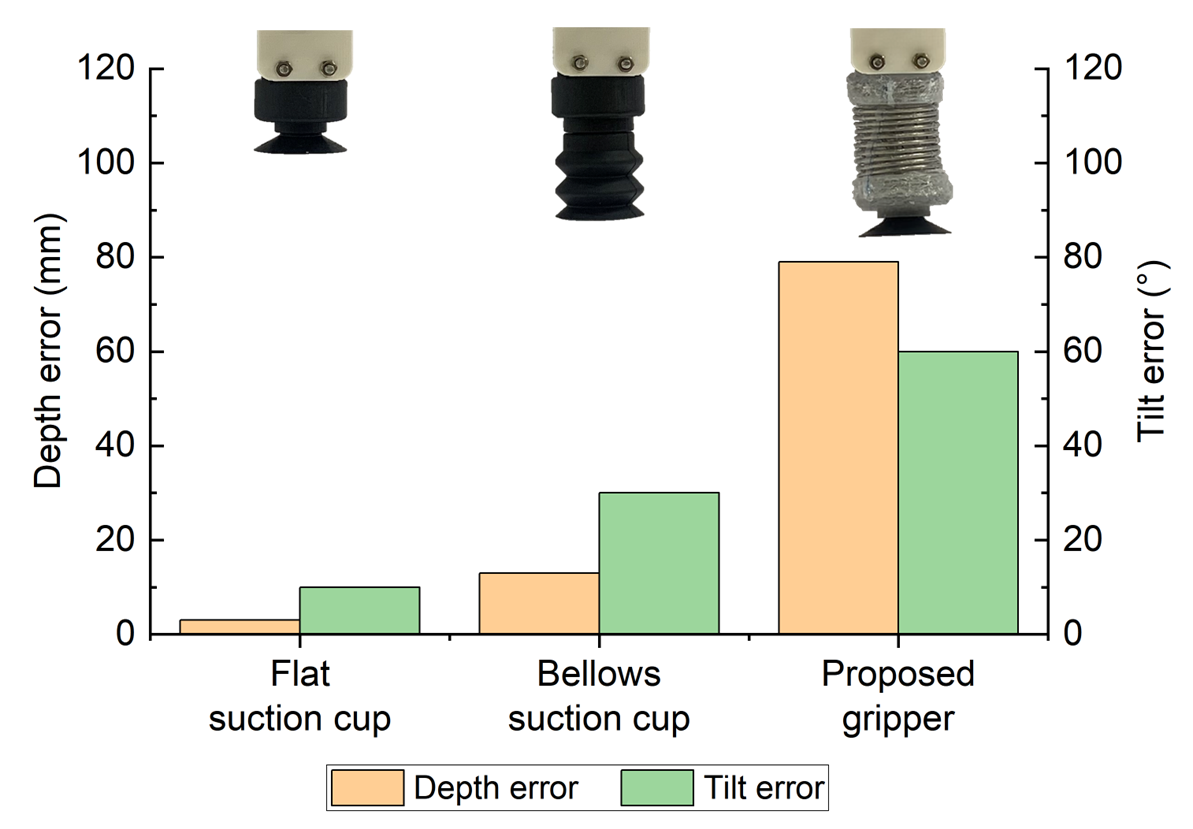}
      \caption{Allowable depth and tilt error of the proposed gripper compared with the flat and bellows suction cups. }
      \label{fig_8}
      \vspace{-0.5cm}
   \end{figure}
The picking force was measured to determine the maximum weight of the object that the gripper could pick according to the picking distance and spring constant. Fig. 9(a) shows the experimental setup. The picking force was measured by lifting the fixed plate attached to the load cell (333FDX, KTOYO). The pressure in the suction cup and gripper body was measured using pressure sensors (ZSE30A-C6H-C, SMC), and the flow rate was fixed at the maximum value of 14 L/min. The gripper was attached to a linear guide such that the distance to the object could be adjusted.

Fig. 9(b) shows the pressure data of the gripper body and the force data measured by the load cell during the picking force experiment. Starting with the contracted gripper (\textcircled{\scriptsize 1} in Fig. 9(b)), the gripper deployed and sealed with a fixed plate, and an upward force was applied to the load cell (\textcircled{\scriptsize 2} in Fig. 9(b)). The picking force was measured as the maximum force in state \textcircled{\scriptsize 2}. To measure the spring constant of the compressible spring built inside the gripper, the restoring force of the spring was measured by connecting the inner and outer chambers of the gripper to the atmosphere (\textcircled{\scriptsize 3} in Fig. 9(b)). The experimental and modeling results for the picking force are shown in Fig. 9(c). The picking force increases for increasing distance or decreasing spring constant, because the restoring force of the spring decreases, and the effective area of the body increases. However, the picking force cannot exceed 45 N ($F_{s,max}$), because the suction cup falls off from the fixed plate. This situation can also be observed in the raw data in Fig. 9(d), where the picking force drops to zero as soon as it reaches $F_{s,max}$. Furthermore, increasing the distance from the point where the suction cup starts to fall off results in a decrease in the picking force, because the suction cup cannot be tightly attached, and the effective area decreases. The picking force model effectively represents the experimental result, although a slight difference occurs due to the buckling of the inner spring or the mass of the gripper, which is not considered in the model.

\subsection{Picking Speed and Reliability}

To test the picking speed, the cycle time of the proposed gripper was measured by using a gripper of wire diameter 1.6 mm. The cycle times for picking a 5 g plate at 65, 95, and 125 mm were analyzed from the video captured at 60 fps. The pressure inside the suction cup and gripper body were additionally measured. The observed cycle time was within 1.2 s for all three distances (Fig. 10(a)). For farther distances, picking required longer time.

Finally, we performed a reliability test (Fig. 10(b)). Picking and placing a thin plate of 5 g was repeated 1000 times with a gripper of wire diameter 1.6 mm. The pressures inside the suction cup and gripper body were measured to monitor the picking state of the gripper. The cycle time of the reliability test was sufficiently set to 3 s, 1.5 s each for deploying and retracting, respectively. The maximum suction cup pressure changed within 4.34 \% during 1000 cycles, but did not affect the gripper performance. The minimum pressure inside the gripper body was maintained at -91 kPa in each cycle, verifying entire sealing and stable picking.

   \begin{figure}[t]
      \centering
      \includegraphics[width=0.97\columnwidth]{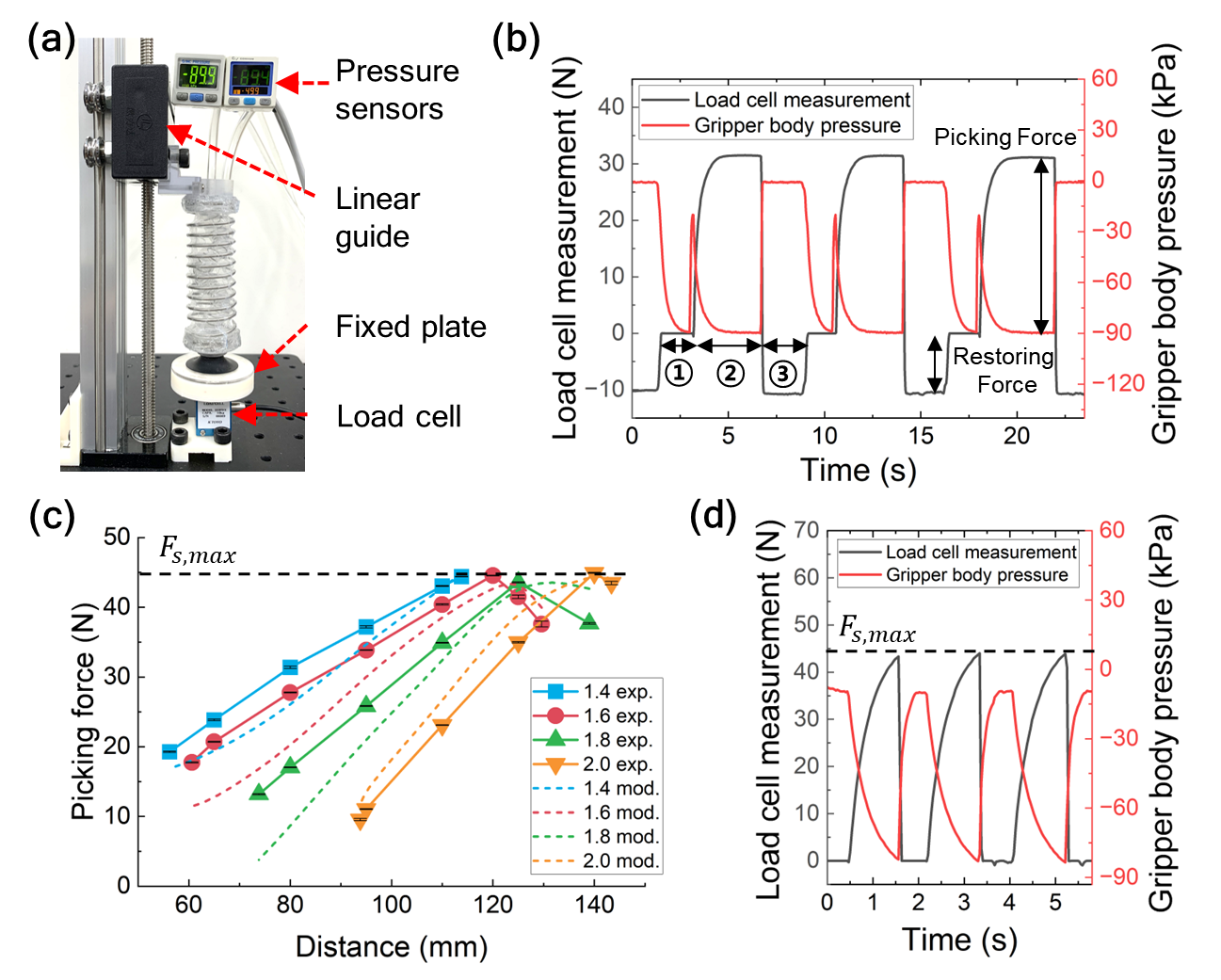}
      \caption{Picking force experiment. (a) Experimental setup for the picking force test. (b) Raw data of the picking force experiment: the gripper is \textcircled{\scriptsize 1} contracted, \textcircled{\scriptsize 2} deploys and pull the fixed plate upwards, and \textcircled{\scriptsize 3} releases it. (c) Picking force according to the distance and spring constant compared with the modeling results. (d) Raw data when the suction cup falls off from the fixed plate. }
      \label{fig_9}
      \vspace{-0.5cm}
   \end{figure}

   \begin{figure}[t]
      \centering
      \includegraphics[width=0.97\columnwidth]{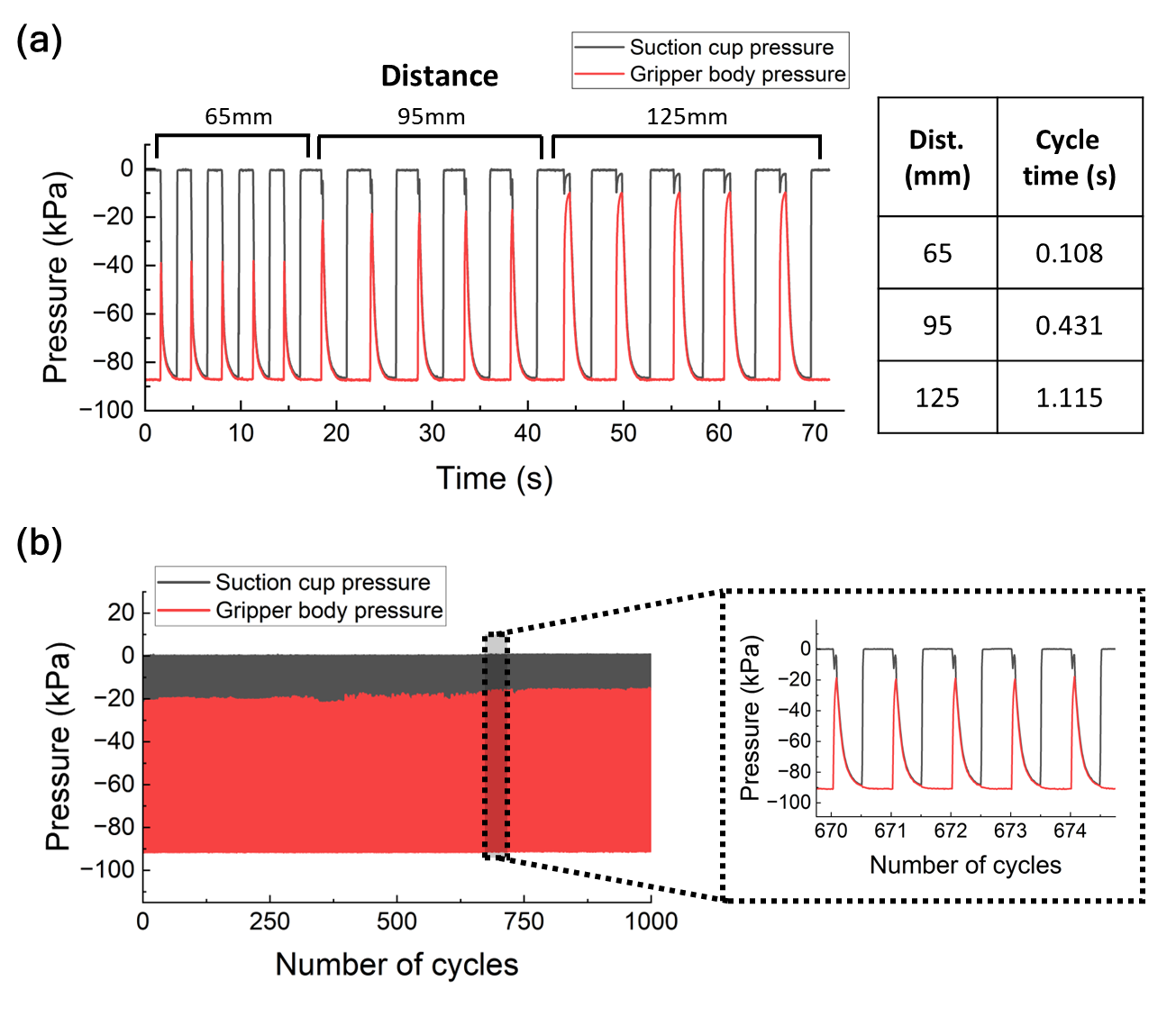}
      \caption{(a) Cycle time experiment according to the distance to the object. (b) Reliability test of 1000 cycles.}
      \label{fig_10}
   \end{figure}

\section{Demonstrations}
To verify the feasibility of the proposed gripper, four demonstrations on the gripper were conducted. The gripper was able to pick objects of different heights from the same picking height (Fig. 11(a)), eliminating the need for depth sensing to pick up objects of different heights. Bin picking of transparent objects is challenging, because vision systems suffer from large depth and tilt errors due to the transparency of objects. The gripper solves the bin-picking of transparent dishes by conforming its body to distant and tilted dishes (Fig. 11(b)). 

Further extensions of the gripper allow the gripper to not only pick up objects against depth and tilt errors but also conduct versatile tasks. By extending the gripper in parallel using multiple grippers, the gripper can pick multiple objects at different heights simultaneously, enabling depalletizing despite the unmatched number of grippers (three) and objects in each layer (four by four)  (Fig. 11(c)). Finally, the extended version of the gripper (140 mm stroke) can penetrate a narrow, deep space that cannot be reached by a manipulator, enabling warehouse picking in occluded environments (Fig. 11(d)).

   \begin{figure}[t]
      \centering
      \includegraphics[width=0.97\columnwidth]{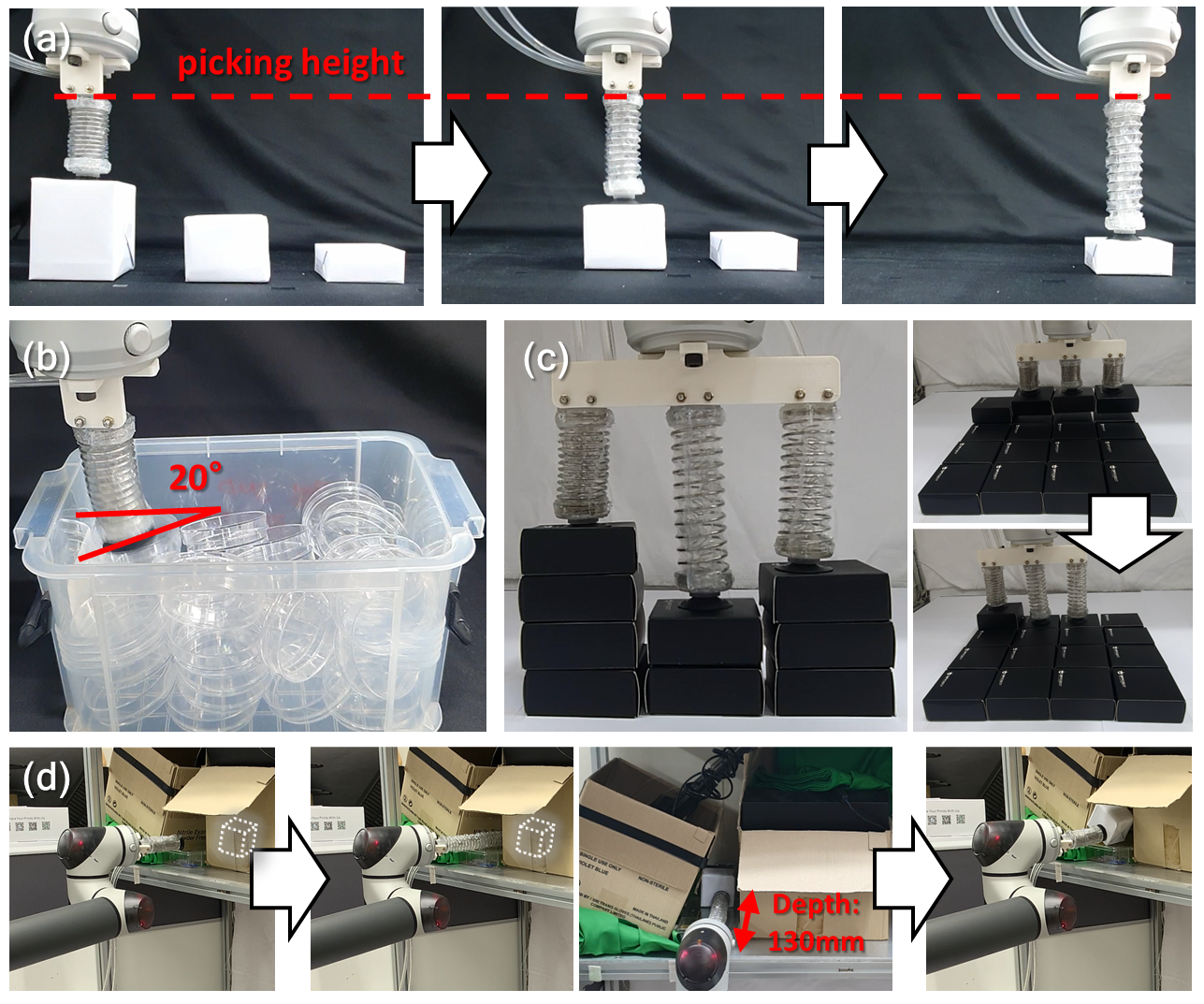}
      \caption{Demonstrations on the proposed gripper. (a) Picking objects with different height from the same picking height. (b) Bin picking of transparent objects that cause large errors in depth sensing. (c) Solving depalletizing task by simultaneously picking objects with different heights. (d) Warehouse picking through a narrow, deep space.}
      \label{fig_11}
      \vspace{-0.5cm}
   \end{figure}

\section{Conclusion}
Our study proposed a longitudinally deployable and initially compact compliant suction gripper that reliably picks objects over wide depth and tilt error ranges. Without additional contact sensing, the proposed gripper can pick objects at various distances and orientations by utilizing a compliant body to conform to the object within the range of deployment. We proposed a gripper body and pneumatic circuit design that enable reliable picking, overcoming depth and tilt errors; the picking by the proposed gripper according to the depth error was analyzed using two analytical models and experimentally validated. In addition, picking angle range, maximum picking force, picking speed, and picking reliability were experimentally measured. The feasibility of picking with depth and tilt errors was verified through two demonstrations and two additional demonstrations were suggested.

Future work may include analytical investigation on tilt angle to optimize the gripper design. Further parametric studies on the thickness or material of the film enclosing the spring would improve the gripper to be resistant to scratches from the external environment. The characteristics of the proposed gripper for robustly picking objects at various distances and angles are expected to be advantageous for mobile platforms such as drones, where the relative position of the gripper and object dynamically changes. Therefore, we expect the technology to be beneficial in static environments with large vision sensing errors and in dynamic situations.

\addtolength{\textheight}{-12cm}   % This command serves to balance the column lengths

%%%%%%%%%%%%%%%%%%%%%%%%%%%%%%%%%%%%%%%%%%%%%%%%%%%%%%%%%%%%%%%%%%%%%%%%%%%%%%%%

%%%%%%%%%%%%%%%%%%%%%%%%%%%%%%%%%%%%%%%%%%%%%%%%%%%%%%%%%%%%%%%%%%%%%%%%%%%%%%%%

\bibliographystyle{IEEEtran}
\bibliography{arXiv-Submission.bib}

\begin{thebibliography}{10}
\providecommand{\url}[1]{#1}
\csname url@samestyle\endcsname
\providecommand{\newblock}{\relax}
\providecommand{\bibinfo}[2]{#2}
\providecommand{\BIBentrySTDinterwordspacing}{\spaceskip=0pt\relax}
\providecommand{\BIBentryALTinterwordstretchfactor}{4}
\providecommand{\BIBentryALTinterwordspacing}{\spaceskip=\fontdimen2\font plus
\BIBentryALTinterwordstretchfactor\fontdimen3\font minus
  \fontdimen4\font\relax}
\providecommand{\BIBforeignlanguage}[2]{{%
\expandafter\ifx\csname l@#1\endcsname\relax
\typeout{** WARNING: IEEEtran.bst: No hyphenation pattern has been}%
\typeout{** loaded for the language `#1'. Using the pattern for}%
\typeout{** the default language instead.}%
\else
\language=\csname l@#1\endcsname
\fi
#2}}
\providecommand{\BIBdecl}{\relax}
\BIBdecl

\bibitem{Pham2019d}
H.~Pham and Q.~C. Pham, ``{Critically fast pick-and-place with suction cups},''
  \emph{Proceedings - IEEE International Conference on Robotics and
  Automation}, vol. 2019-May, pp. 3045--3051, 2019.

\bibitem{Papadakis2020c}
E.~Papadakis, F.~Raptopoulos, M.~Koskinopoulou, and M.~Maniadakis, ``{On the
  Use of Vacuum Technology for Applied Robotic Systems},'' \emph{2020 6th
  International Conference on Mechatronics and Robotics Engineering, ICMRE
  2020}, pp. 73--77, 2020.

\bibitem{Huh2021c}
T.~M. Huh, K.~Sanders, M.~Danielczuk, M.~Li, Y.~Chen, K.~Goldberg, and H.~S.
  Stuart, ``{A Multi-Chamber Smart Suction Cup for Adaptive Gripping and Haptic
  Exploration},'' \emph{IEEE International Conference on Intelligent Robots and
  Systems}, pp. 1786--1793, 2021.

\bibitem{Mueller2016a}
V.~Mueller, R.~Behrens, and N.~Elkmann, ``{A new multi-modal approach towards
  reliable bin-picking application},'' \emph{47th International Symposium on
  Robotics, ISR 2016}, vol. 2016, pp. 473--479, 2016.

\bibitem{Schwarz2017}
M.~Schwarz, A.~Milan, C.~Lenz, A.~Munoz, A.~S. Periyasamy, M.~Schreiber,
  S.~Schuller, and S.~Behnke, ``{NimbRo picking: Versatile part handling for
  warehouse automation},'' \emph{Proceedings - IEEE International Conference on
  Robotics and Automation}, vol.~0, pp. 3032--3039, 2017.

\bibitem{Joffe2019c}
B.~Joffe, T.~Walker, R.~Gourdon, and K.~Ahlin, ``{Pose estimation and bin
  picking for deformable products},'' \emph{IFAC-PapersOnLine}, vol.~52,
  no.~30, pp. 361--366, 2019.

\bibitem{Burks2005a}
T.~Burks, F.~Villegas, M.~Hannan, S.~Flood, B.~Sivaraman, V.~Subramanian, and
  J.~Sikes, ``{Engineering and horticultural aspects of robotic fruit
  harvesting: Opportunities and constraints},'' \emph{HortTechnology}, vol.~15,
  no.~1, pp. 79--87, 2005.

\bibitem{Sajjan2020a}
S.~Sajjan, M.~Moore, M.~Pan, G.~Nagaraja, J.~Lee, A.~Zeng, and S.~Song,
  ``{Clear Grasp: 3D Shape Estimation of Transparent Objects for
  Manipulation},'' \emph{Proceedings - IEEE International Conference on
  Robotics and Automation}, pp. 3634--3642, 2020.

\bibitem{Baek2020}
E.-T. Baek, H.-J. Yang, S.-H. Kim, G.~Lee, and H.~Jeong, ``Distance error
  correction in time-of-flight cameras using asynchronous integration time,''
  \emph{Sensors}, vol.~20, no.~4, 2020.

\bibitem{Kim2017c}
K.~Kim and H.~Shim, ``{Robust approach to reconstructing transparent objects
  using a time-of-flight depth camera},'' \emph{Optics Express}, vol.~25,
  no.~3, p. 2666, 2017.

\bibitem{Chai2020e}
C.~Y. Chai, Y.~P. Wu, and S.~L. Tsao, ``{Deep Depth Fusion for Black,
  Transparent, Reflective and Texture-Less Objects},'' \emph{Proceedings - IEEE
  International Conference on Robotics and Automation}, pp. 6766--6772, 2020.

\bibitem{Piao2020d}
Y.~Piao, Z.~Rong, M.~Zhang, W.~Ren, and H.~Lu, ``{A2dele: Adaptive and
  attentive depth distiller for efficient RGB-D salient object detection},''
  \emph{Proceedings of the IEEE Computer Society Conference on Computer Vision
  and Pattern Recognition}, vol.~2, pp. 9057--9066, 2020.

\bibitem{Wang2021d}
K.~Wang, Z.~Zhang, Z.~Yan, X.~Li, B.~Xu, J.~Li, and J.~Yang, ``{Regularizing
  Nighttime Weirdness: Efficient Self-supervised Monocular Depth Estimation in
  the Dark},'' \emph{Proceedings of the IEEE International Conference on
  Computer Vision}, pp. 16\,035--16\,044, 2021.

\bibitem{Aoyagi2020c}
S.~Aoyagi, M.~Suzuki, T.~Morita, T.~Takahashi, and H.~Takise, ``{Bellows
  Suction Cup Equipped with Force Sensing Ability by Direct Coating Thin-Film
  Resistor for Vacuum Type Robotic Hand},'' \emph{IEEE/ASME Transactions on
  Mechatronics}, vol.~25, no.~5, pp. 2501--2512, 2020.

\bibitem{Koyama2019d}
K.~Koyama, M.~Shimojo, A.~Ming, and M.~Ishikawa, ``{Integrated control of a
  multiple-degree-of-freedom hand and arm using a reactive architecture based
  on high-speed proximity sensing},'' \emph{International Journal of Robotics
  Research}, vol.~38, no.~14, pp. 1717--1750, 2019.

\bibitem{Baek2021c}
S.~Baek and D.~O. Kim, ``{Predictive process adjustment by detecting system
  status of vacuum gripper in real time during pick-up operations},''
  \emph{Processes}, vol.~9, no.~4, 2021.

\bibitem{Liu2020b}
S.~Liu, W.~Dong, Z.~Ma, and X.~Sheng, ``{Adaptive Aerial Grasping and Perching
  with Dual Elasticity Combined Suction Cup},'' \emph{IEEE Robotics and
  Automation Letters}, vol.~5, no.~3, pp. 4766--4773, 2020.

\bibitem{Cui2013a}
Y.~Cui, S.~Schuon, S.~Thrun, D.~Stricker, and C.~Theobalt, ``{Algorithms for 3D
  shape scanning with a depth camera},'' \emph{IEEE Transactions on Pattern
  Analysis and Machine Intelligence}, vol.~35, no.~5, pp. 1039--1050, 2013.

\bibitem{Chang2021a}
C.~C. Chang, K.~C. Shih, H.~C. Ting, and Y.~S. Su, ``{Utilizing Machine
  Learning to Improve the Distance Information from Depth Camera},''
  \emph{Proceedings of the 3rd IEEE Eurasia Conference on IOT, Communication
  and Engineering 2021, ECICE 2021}, pp. 405--408, 2021.

\bibitem{Doi2020c}
S.~Doi, H.~Koga, T.~Seki, and Y.~Okuno, ``{Novel Proximity Sensor for Realizing
  Tactile Sense in Suction Cups},'' \emph{Proceedings - IEEE International
  Conference on Robotics and Automation}, pp. 638--643, 2020.

\bibitem{Sitti2021d}
M.~Sitti, ``{Physical intelligence as a new paradigm},'' \emph{Extreme
  Mechanics Letters}, vol.~46, p. 101340, 2021.

\bibitem{Laschi2016a}
C.~Laschi, B.~Mazzolai, and M.~Cianchetti, ``{Soft robotics: Technologies and
  systems pushing the boundaries of robot abilities},'' \emph{Science
  Robotics}, vol.~1, no.~1, pp. 1--12, 2016.

\bibitem{Li2019}
S.~Li, J.~J. Stampfli, H.~J. Xu, E.~Malkin, E.~V. Diaz, D.~Rus, and R.~J. Wood,
  ``{A Vacuum-driven Origami “ Magic-ball ” Soft Gripper},'' pp.
  7401--7408, 2019.

\bibitem{Amend2012g}
J.~R. Amend, E.~Brown, N.~Rodenberg, H.~M. Jaeger, and H.~Lipson, ``{A positive
  pressure universal gripper based on the jamming of granular material},''
  \emph{IEEE Transactions on Robotics}, vol.~28, no.~2, pp. 341--350, 2012.

\bibitem{Zhou2017c}
J.~Zhou, S.~Chen, and Z.~Wang, ``{A Soft-Robotic Gripper with Enhanced Object
  Adaptation and Grasping Reliability},'' \emph{IEEE Robotics and Automation
  Letters}, vol.~2, no.~4, pp. 2287--2293, 2017.

\bibitem{Pounds2011d}
P.~E. Pounds, D.~R. Bersak, and A.~M. Dollar, ``{Practical aerial grasping of
  unstructured objects},'' \emph{2011 IEEE Conference on Technologies for
  Practical Robot Applications, TePRA 2011}, pp. 99--104, 2011.

\bibitem{Angelini2020d}
F.~Angelini, C.~Petrocelli, M.~G. Catalano, M.~Garabini, G.~Grioli, and
  A.~Bicchi, ``{SoftHandler: An Integrated Soft Robotic System for Handling
  Heterogeneous Objects},'' \emph{IEEE Robotics and Automation Magazine},
  vol.~27, no.~3, pp. 55--72, 2020.

\bibitem{Vac-cyl}
D.~Pascoe, ``{Vacuum Pick-Up Cylinders},'' fluidpowerjournal.com.
  https://fluidpowerjournal.com/vacuum-pick-up-cylinders/ (accessed Sep. 7,
  2022).

\bibitem{Liu2020}
S.~Liu, W.~Dong, Z.~Ma, and X.~Sheng, ``{Adaptive Aerial Grasping and Perching
  with Dual Elasticity Combined Suction Cup},'' \emph{IEEE Robotics and
  Automation Letters}, vol.~5, no.~3, pp. 4766--4773, 2020.

\bibitem{McCulloch2016}
T.~McCulloch and D.~Herath, ``{Towards a universal gripper - On the use of
  suction in the context of the Amazon Robotic Challenge},'' \emph{Australasian
  Conference on Robotics and Automation, ACRA}, vol. 2016-December, pp.
  153--160, 2016.

\bibitem{FluidMech}
F.~M. White, \emph{{Fluid Mechanics}}.\hskip 1em plus 0.5em minus 0.4em\relax
  7th ed. New York, NY: Mc Graw Hill, 2011, pp.362--381.

\bibitem{Lee2019a}
J.~G. Lee and H.~Rodrigue, ``{Origami-Based Vacuum Pneumatic Artificial Muscles
  with Large Contraction Ratios},'' \emph{Soft Robotics}, vol.~6, no.~1, pp.
  109--117, 2019.

\end{thebibliography}

\end{document}